\documentclass[10pt,twocolumn,letterpaper]{article}

\usepackage{cvpr}
\usepackage{times}
\usepackage{epsfig}
\usepackage{graphicx}
\usepackage{amsmath}
\usepackage{amssymb}

\usepackage{booktabs}
\usepackage{multicol}
\usepackage{multirow}
\usepackage[ruled,vlined]{algorithm2e}
\usepackage{soul,xcolor}
\usepackage{pifont}

\newcommand*{\affaddr}[1]{#1} 
\newcommand*{\affmark}[1][*]{\textsuperscript{#1}}
\newcommand*{\email}[1]{\texttt{#1}}

\usepackage[breaklinks=true,bookmarks=false]{hyperref}

\cvprfinalcopy 

\def\cvprPaperID{****} 

\begin{document}

\title{AutoPose: Searching Multi-Scale Branch Aggregation for Pose Estimation}


\author{%
$\text{Xinyu Gong}^{*}$\affmark[1] \quad $\text{Wuyang Chen}^{*}$\affmark[1] \quad Yifan Jiang\affmark[1] \quad Ye Yuan\affmark[2] \\ Xianming Liu\affmark[3] \quad Qian Zhang\affmark[3] \quad Yuan Li\affmark[3] \quad Zhangyang Wang\affmark[1]\\
\affaddr{ \affmark[1] Department of Electrical and Computer Engineering, The University of Texas at Austin \\
\affmark[2]Department of Computer Science \& Engineering, Texas A\&M University \, \affmark[3]Horizon Robotics Inc.}
\\
\email{\{xinyu.gong, wuyang.chen, yifanjiang97, atlaswang\}@utexas.edu} \\ \email{ye.yuan@tamu.edu}, \, \email{\{xianming.liu, qian01.zhang, yuan.li\}@horizon.ai}\\
}

\maketitle
\renewcommand{\thefootnote}{\fnsymbol{footnote}}
\footnotetext[1]{Part of this work was done while Xinyu and Wuyang were Research Interns at Horizon Robotics Applied AI Lab.}

\begin{abstract}
    We present AutoPose, a novel neural architecture search (NAS) framework that is capable of automatically discovering multiple parallel branches of cross-scale connections towards accurate and high-resolution 2D human pose estimation. Recently, high-performance hand-crafted convolutional networks for pose estimation show growing demands on multi-scale fusion and high-resolution representations. However, current NAS works exhibit limited flexibility on scale searching, they dominantly adopt simplified search spaces of single-branch architectures. Such simplification limits the fusion of information at different scales and fails to maintain high-resolution representations. The presented AutoPose framework is able to search for multi-branch scales and network depth, in addition to the cell-level micro structure. Motivated by the search space, a novel bi-level optimization method is presented, where the network-level architecture is searched via reinforcement learning, and the cell-level search is conducted by the gradient-based method. Within 2.5 GPU days, AutoPose is able to find very competitive architectures on the MS COCO dataset, that are also transferable to the MPII dataset. Our code is available at \href{https://github.com/VITA-Group/AutoPose}{https://github.com/VITA-Group/AutoPose}.
    
\end{abstract}

\vspace{-10pt}
\section{Introduction}
\vspace{-4pt}
Human 2D pose estimation targets at localizing anatomical keypoints of individuals from images. Accurate pose estimation plays an essential role in understanding human behaviors. This paper focuses on pose estimation for single person, which can be a cornerstone for downstream tasks including action recognition \cite{rodriguez2008action,ryoo2009spatio,schuldt2004recognizing}, gaming \cite{ke2010real}, multi-person pose estimation \cite{cao2017realtime}, and video tracking \cite{raaj2019efficient,xiu2018pose}, etc.

\begin{table*}[h]
\centering
\caption{Comparison of AutoPose against other NAS works. AutoPose is the first NAS framework that integrally searches for cell, scale, aggregation (``Agg.'') and branches. We also compares the size of search space (``SS''), optimization method, and search time (GPU days).} \vspace{0.5em}. 
{\small
\begin{tabular}{l|ccccccccc}
\toprule
Model & SS & Optimization & Cell & Scale & Agg. & Branch & Dataset & Task & Days \\ \midrule
NASNet \cite{zoph2018learning} & $10^{16}$ & RL & \checkmark &  &  &  & CIFAR-10& Cls & 2000 \\
DARTS \cite{liu2018darts} & $10^{28}$ & GRAD & \checkmark &  &  & & CIFAR-10 & Cls & 4 \\
DPC \cite{chen2018searching} & $10^{11}$ & RS &  &  & \checkmark &  & Cityscapes & Seg & 2600 \\
Auto-DeepLab \cite{liu2019auto} & $10^{19}$ & GRAD & \checkmark & \checkmark &  & & Cityscapes & Seg & 3 \\ \hline
Ours & $10^{61}$ & GRAD+RL & \checkmark & \checkmark & \checkmark & \checkmark & MSCOCO & Pose & 2.5 \\ \bottomrule
\end{tabular}
}\label{table:comparison}
\end{table*}

Recently, great endeavors have been made to exploit convolutional neural networks (CNN) to improve pose estimation. Especially, multi-scale branches and context aggregation modules are increasingly leveraged to capture multi-grained features. For example, Newell \etal proposes the stacked Hourglass network \cite{newell2016stacked}, which repeatedly downsamples and restores resolutions through the network.
Cascaded pyramid network \cite{chen2018cascaded} constructs a multi-scale network, keeping multi-scale information through the network and aggregating them at the end.
HRNet \cite{sun2019deep} achieves state-of-the-art performance by constructing a multi-branch network, where high-resolution representations are maintained through the whole pipeline, in addition to multi-scale aggregation at each stage. While multi-scale features, as well as high resolution, are shown to be vital for accurate pose estimation, how to advance the design of pose estimation networks further is not immediately clear, since the arising complex branches and connectivity patterns can cost heavy human efforts to explore thoroughly.

Designing modern CNN architectures manually can result in tedious trail-and-errors. To this end, the neural architecture search (NAS) has recently drawn a booming interest. Aiming to discover an optimal network architecture from data, NAS has been successfully applied primarily on image classification \cite{dong2019searching,liu2018darts,negrinho2017deeparchitect,pham2018efficient,real2019regularized,xie2018snas,zoph2016neural}, and lately also on object detection \cite{chen2019detnas,ghiasi2019fpn}, semantic segmentation \cite{liu2019auto,nekrasov2019fast,chen2019fasterseg}, person re-identification \cite{quan2019auto}, speech recognition \cite{ding2020autospeech}, super-resolution \cite{song2020efficient}, medical image analysis \cite{zhu2019v}, and even generative models \cite{gong2019autogan,fu2020autogan} or Bayesian deep networks \cite{ardywibowo2020nads}. These works surge heavy interests in designing sophisticated architectures automatically, reducing human efforts as well as revealing/verifying design insights.

Despite the promise of NAS, there remains to be a major gap between its current capability, and the real sophistication needed by advanced computer vision applications. For example, for the sake of simplicity, most NAS methods usually define their search spaces in a way that can only cover chain-like single backbones (with skip connections), which will exclude the discovery of multiple (and inter-connected) backbones that have proven effectiveness \cite{chen2018cascaded,sun2019deep}. Recently, Auto-DeepLab \cite{liu2019auto} proposes a two-level hierarchy search space for semantic segmentation, which is able to search for the network scales at different stages. However, considering only connections within a single cell, Auto-DeepLab cannot derive an architecture that contains connections over multiple sub-paths over different scales at the same time. 
NAS-FPN \cite{ghiasi2019fpn} designs cross-scale connections only for the feature fusion head, while the backbone is not optimized.

This paper proposes AutoPose, a novel NAS framework that automatically designs architectures towards accurate 2D human pose estimation. The core of  AutoPose is a hierarchical multi-scale \textbf{search space}, consisting of a novel network-level search space in addition to the cell-level search space. Our search space can simultaneously maintain multiple parallel branches of diverse scales in an end-to-end fashion throughout the network, which leads to much flexible and finer-grained exploration. To effectively fuse features from branches, AutoPose further searches for novel aggregation cells, which support dense and sophisticated cross-scale aggregations, instead of naively summing up all features \cite{sun2019deep}. 

As a result, the AutoPose search space is unusually gigantic ($10^{61}$ candidates) with multi-level selections (from cell to network), making either gradient method or reinforcement learning alone \textit{not directly applicable}. For example, traditional gradient-based search is unable to select more than one branches simultaneously \cite{liu2019auto}. On the other hand, reinforcement learning would be too time-consuming for such a huge space. Motivated by that, we present a novel bi-level \textbf{optimization algorithm} combining the two, for different aspects of search space.  Gradient-based method is leveraged to search for the micro cell architectures (cell-level). An RNN controller trained via reinforcement learning then guides the dynamic searching over multiple scales and network depth (network level), which boosts the flexibility in discovering networks of capacity variations.

We summarize our contributions as follows: \vspace{-2pt}
\begin{itemize}
    \item \textbf{High-resolution Multi-branch Search Space.} We show that: i) the searchable parallel branches of multiple scales; and ii) the cross-scale context aggregations, are two key components towards searching for accurate human pose estimation.
    \vspace{-1pt}
    Extensive ablation studies support our design philosophy.
    \item \textbf{Bi-level Optimization.} To fulfill the effective discovery over the gigantic fined-grained search space, we integrate the reinforcement learning and the gradient-based search method for the first time, to jointly optimize the network-level and the cell-level architectures.\vspace{-1pt}
    \item Experiments demonstrate that AutoPose for the first time automatically designs architectures of sophisticated connections, achieving competitive performance as hand-crafted networks for human pose estimation.\vspace{-2pt}
\end{itemize}

We noticed another recently proposed NAS framework of pose estimation called PoseNFS \cite{yang2019pose}. Unlike \cite{liu2019auto}, PoseNFS follows a one-together search strategy directly over macro and micro levels. To simplify their search, the authors referred to the body structure prior knowledge to decompose the whole search space into multiple part-specific ones, with global pose constraint relationship enforced. However, its performance falls behind current state-of-the-art hand-crafted pose estimation models with a notable margin. In comparison, we decouple our search at two different granularity levels in one unified space, and show the feasibility and effectiveness of purely end-to-end data-driven search without drawing any explicit body part knowledge. We compare our results with PoseNFS in Section 4.

\vspace{-3pt}
\section{Related Work}
\vspace{-2pt}
\subsection{Human Pose Estimation}
\vspace{-2pt}

Existing pose estimation work can be categorized to either bottom-up or top-down. Bottom-up methods \cite{cao2017realtime,insafutdinov2016deepercut} first detect all parts in the image (i.e. parts of every person), then associate or group parts belonging to each individual. Bottom-up methods are more suitable for efficient pose estimation. Top-down methods~\cite{chen2018cascaded,fang2017rmpe,he2017mask,huang2017coarse,papandreou2017towards,sun2019deep} first generate human proposals in the form of bounding boxes and then exploit non-maximum suppression (NMS) to remove the redundant proposals, after which a single-person pose estimation is conducted for every proposal.

The recent trend in pose estimation leverages high-resolution features and aggregating contexts from multiple scales. Chen \etal~\cite{chen2018cascaded} proposes the Cascaded Pyramid Network which consists of a GlobalNet based on feature pyramid and a cascaded multi-scale RefineNet to focus more on fine-granularity hard keypoints. HRNet \cite{sun2019deep} achieves state-of-the-art performance by utilizing a network of multiple branches across multiple scales, where information is aggregated at each stage of the network. In AutoPose, we follow this line of leveraging high-resolution features across multi-scale branches from end to end.

\begin{figure*}[ht!]
\begin{center}
    \vspace{-4pt}
 \includegraphics[width=0.8\linewidth]{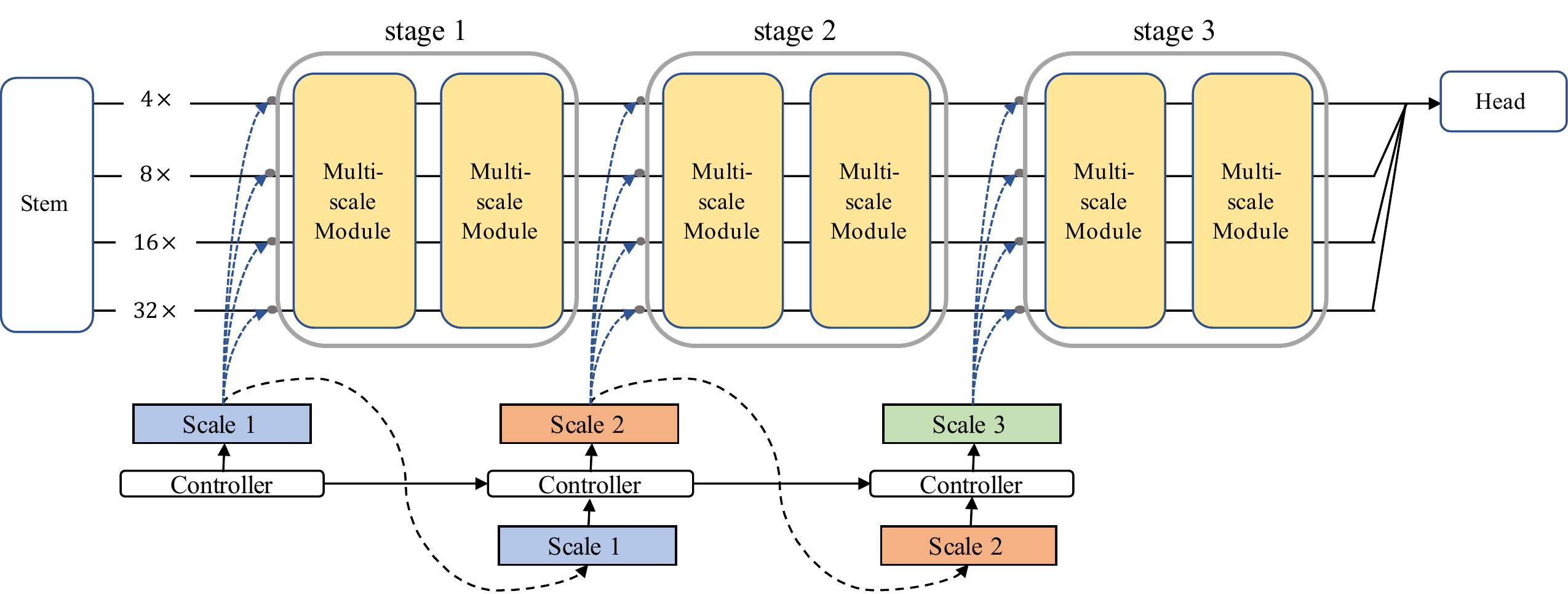}
\end{center}
\vspace{-1em}
   \caption{Our AutoPose framework. We support the multi-scale parallel branches and context aggregation via our stacked multi-scale modules. The activation of each scale ($\{4\times, 8\times, 16\times, 32\times\}$ when $B=4$ shown here) in each stage is controlled by an RNN controller.
    }
    \vspace{-4pt}
\label{fig:macro}
\end{figure*}

\vspace{-2pt}
\subsection{Neural Architecture Search Spaces}
\vspace{-2pt}
Most existing works in neural architecture search (NAS) focus on searching either stacked operators \cite{pham2018efficient,zoph2016neural} or repeated cell-structured directed acyclic graph \cite{liu2018darts} for the classification task. Search spaces in these works are mainly designed for chain-like single-path networks with fixed scale patterns. Recent NAS works explore more challenging high-level vision tasks like object detection \cite{chen2019detnas,ghiasi2019fpn} and semantic segmentation \cite{liu2019auto,nekrasov2019fast}. Although these dense prediction tasks are more sensitive to resolutions and require aggregation of multi-scale context for improving performance, these early NAS works for dense prediction tasks are still limited in their searchable scale-level connections. In contrast, our search space supports more complex cross-scale connections throughout the network. Different from the network-level search in Auto-Deeplab \cite{liu2019auto} which still only explore and derive only one branch, our AutoPose supports maintaining multiple parallel branches across scales. Our search space naturally provides the architecture exploration with more flexibility yet it is more challenging.

\vspace{-2pt}
\subsection{Neural Architecture Search Methods}
\vspace{-2pt}
Neural architecture optimization is first proposed with the reinforcement learning method \cite{pham2018efficient,zoph2016neural}, which leverages a learned policy to control the selection of operators along with the network. Reinforcement learning is a popular methods for NAS works which involve non-differentiable optimization target \cite{gong2019autogan,tan2019mnasnet}, as the policy-gradient method \cite{williams1992simple} can leverage discrete extrinsic rewards. The gradient-based method is proposed in \cite{liu2018darts} which provides an efficient way of searching the architecture using gradient descent \cite{liu2019auto}. Real \etal \cite{real2019regularized} successfully introduce an evolution algorithm into NAS, achieving competitive performance, also followed by many \cite{yang2020cars}. Egrinho \etal \cite{negrinho2017deeparchitect} takes the first step to incorporate the Monte Carlo tree search (MCTS) into NAS. \cite{wang2020dc} also adopted the search space divide-and-conquer strategy. NAO \cite{luo2018neural} embeds the complex network structures into a latent space to simplify the searching. All existing NAS works focus on using one search method coupled with their search spaces. However, a single search method may lack the capability of supporting complex search spaces. In our AutoPose, we for the first time integrate different methods for different parts of our search space.
\vspace{-4pt}



\begin{figure}[t]
\begin{center}
 \includegraphics[width=0.85\linewidth]{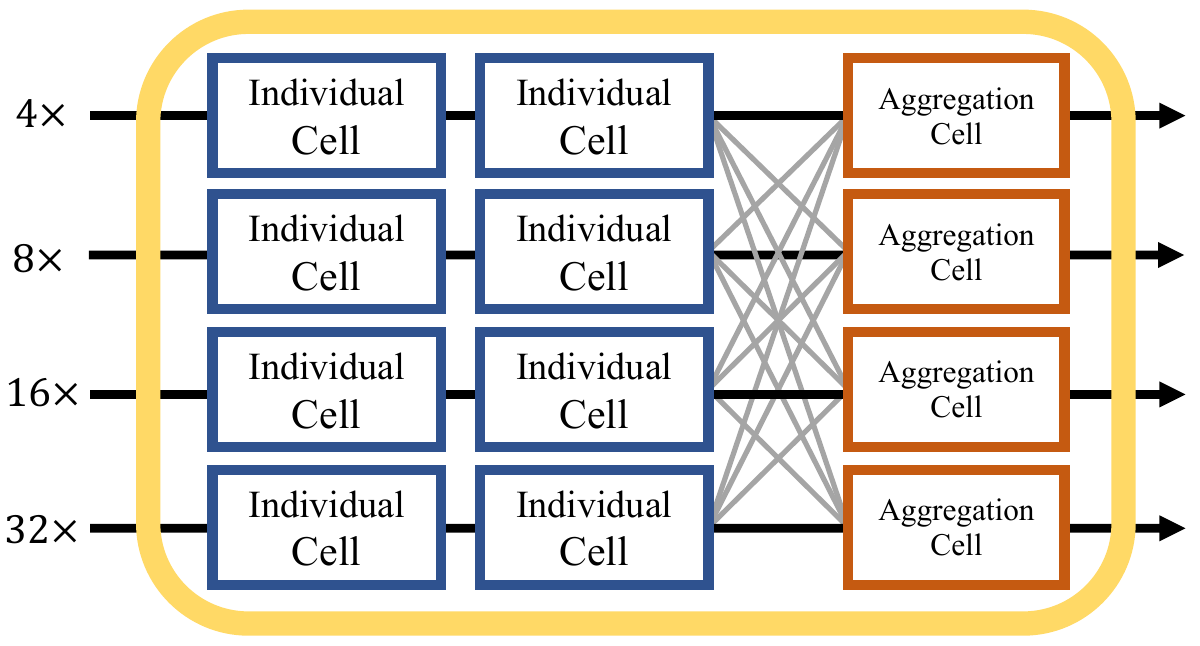}
\end{center}
\vspace{-1em}
   \caption{Our multi-scale module takes inputs of different scales. The inputs will first go through two individual cells and then will be aggregated together by the aggregation cells.}
\label{fig:module}
\vspace{-8pt}
\end{figure}

\section{Method}
\vspace{-4pt}
We illustrate the overview of our AutoPose framework in Fig. \ref{fig:macro}.
The input image is first passed through a pre-defined stem module (Sec. \ref{sec:stem_head}) to output feature maps of different scales\footnote{A feature map of scale $s$ is of $\frac{1}{s^2}$ size of the input image by scaling both the height and width to $\frac{1}{s}$.}, which are served as inputs into our multi-branch search space. We split our search space into three stages, where the activated parallel branches (of different scales) in each stage are individually controlled by our RNN controller. The purpose of these separate stages is to support the different needs of scales from early to late stage along with the network. At the end, the outputs of different branches are upsampled via nearest-neighbor interpolation to the highest resolution, and are aggregated by summation.

\begin{figure}[h]
\begin{center}
 \includegraphics[width=1.0\linewidth]{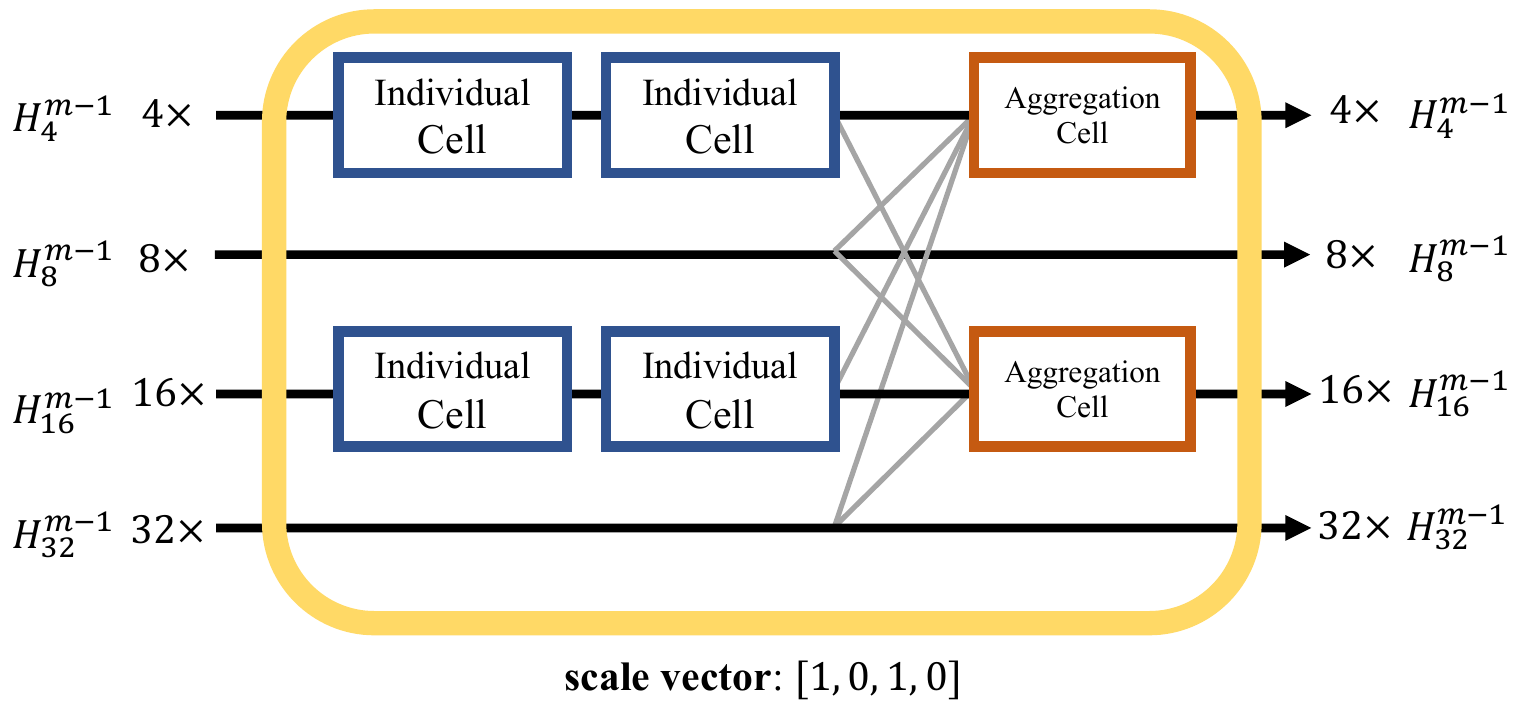}
\end{center}
\vspace{-1em}
   \caption{An example of a multi-scale module with a scale vector $[1, 0, 1, 0]$. Only the branches of scales $4 \times$ and $16 \times$ are activated.}
\label{fig:example}
 \vspace{-10pt}
 \end{figure}

\vspace{-2pt}
\subsection{High-resolution Multi-branch Search Space}
\vspace{-2pt}
Parallel branches of diverse scales and context aggregation modules are two key points for sufficiently capturing feature maps under different receptive fields, which motivates the design of our core functional unit ``Multi-scale Module'' (Fig. \ref{fig:module}). In the following sections, we describe our network-level search space for maintaining multiple parallel branches throughout the network and cell-level structures for merging multi-scale features.
We by default stack two identical multi-scale modules in each stage.

\vspace{-8pt}
\subsubsection{Network-level Search Space}
\vspace{-4pt}
Our core component ``Multi-scale Module'' (Fig. \ref{fig:module}) is composted of $B$ branches of different scales, where each branch includes two individual cells ($\mathcal{C}_{I1}, \mathcal{C}_{I2}$) and one aggregation cell ($\mathcal{C}_M$) (see Sec. \ref{sec:cell}). The activation status of each scale in each stage is individually controlled by our RNN controller. Specifically, each stage may activate one or more different scales. If a scale is activated, the input tensor of that scale will be processed by the two individual cells on that branch. Otherwise, the input tensor will go through a skip connection, which means the two individual cells and the aggregation cell on this deactivated branch will not participate in the network forward and backward process, and correspondingly the cell weights and architecture will not be updated by the gradient descent. The feature maps remain the same scale within each branch.

More formally, for the $m$-th multi-scale module, the RNN controller outputs a binary scale vector $[\beta^m_{2^2}, \beta^m_{2^3}.., \beta^m_{2^{B+1}}]$. For the $b$-th branch (top to bottom in Fig. \ref{fig:macro}, $b \in N^+, 1 \leq b \leq B$) whose scale $s_b = 2^{b+1}$, its output, denoted as $H^m_{s_b}$, can be formulated as:
\vspace{-6pt}
\begin{equation}
    H^m_{s_b} = 
    \begin{cases}
    \mathcal{C}^m_{M,s_b} \circ \mathcal{C}^m_{I2,s_b} \circ \mathcal{C}^m_{I1,s_b} (H^{m-1}_{s_b}) & \text{if $\beta^m_{s_b} = 1$,} \\
    H^{m-1}_{s_b} & \text{if $\beta^m_{s_b} = 0$.}
    \end{cases}
\end{equation}
Note that we allow the situation that all branches in a stage are deactivated, which means that our framework is able to search for the \textbf{network depth}. Fig. \ref{fig:example} shows an example when $B = 4$ and the controller generates a scale vector $[1, 0, 1, 0]$. In this scenario, only the branches of $4\times$ and $16\times$ scales are activated and will process their input tensors, while the branches of $8\times$ and $32\times$ scales become skip connections. Note that the aggregation cells of $4\times$ and $16\times$ scales will take $\mathcal{C}^m_{I2,s_b} \circ \mathcal{C}^m_{I1,s_b} (H^{m-1}_{s_b})$ as inputs for $s_b = 4, 16$, and will take $H^{m-1}_s$ as inputs for $s_b = 8, 32$.

\vspace{-8pt}
\subsubsection{Cell-level Search Space} \label{sec:cell}
\vspace{-4pt}
We include two types of cells ((network basic units): the individual cell and the aggregation cell. As in Fig. \ref{fig:module}, on each branch in the multi-scale module, two individual cells ($\mathcal{C}_{I1}, \mathcal{C}_{I2}$) are sequentially stacked, followed by a aggregation cell ($\mathcal{C}_M$). Both the individual cell $\mathcal{C}_I$ and the aggregation cell $\mathcal{C}_M$ are directed acyclic graphs (DAG), consisting of an ordered sequence of $K_I, K_M$ nodes, respectively. Each node  $x^{(i)}$ is a feature map and each directed edge $(i, j)$ is associated with a specific operation $O^{(i,j)}$ that transforms node. The output of a cell, denoted as $H$, is the concatenation of all nodes, followed by a $1 \times 1$ convolution, aiming at compressing the channel dimension.

\vspace{-8pt}
\paragraph{Individual Cell}
An individual cell $\mathcal{C}_I$ takes its predecessor (the previous cell's output) as its single input, and outputs one tensor.
The $i$-th node $x^{(i)}_I$ is calculated as the sum of the transformed (processed by an operation)  $i$ previous nodes and the input of the cell. 
The computation of $i$-th intermediate node in the individual cell is formulated as:
\begin{equation}
    x^{(i)}_I = \sum_{j<i}O^{(i,j)}(x^{(j)}_I).
    \label{eq:node}
\end{equation}
$O^{(i,j)}$ denotes the operation of $i$-th node, which takes the $j$-th node $x^{(j)}_I$ as input.

\vspace{-8pt}
\paragraph{Aggregation Cell}
A aggregation cell $\mathcal{C}_M$ takes $B$ outputs of different scales from $B$ individual cells as its inputs.
In particular, the $B$ inputs from the individual cells will be first downsampled or upsampled accordingly to match the current aggregation cell's scale.
The computation at $i$-th intermediate node in the aggregation cell is formulated as:
\begin{equation}
    x^{(i)}_M = \sum_{j<i}O^{(i,j)}(x^{(j)}_M) + \sum_{b=1}^B O^i_b(H_{I2, s_b}),
    \label{eq:node}
\end{equation}
where $H_{I2, s_b}$ denotes the output of the second individual cell of scale $s_b = 2^{b+1}$ ($b \in N^+, 1 \leq b \leq B$), with associated operation $O^i_b$ for the $i$-th node.

\vspace{-8pt}
\paragraph{Continuous Relaxation of Cell-level Search Space}
We adopts the continuous relaxation \cite{dong2019searching,liu2019auto,liu2018darts} to represent the cell-level architecture. Specifically, the categorical choice of a particular operation $O^{(i,j)}$ is relaxed to be continuous:
\begin{equation}
    \overline{O}^{(i,j)}(x^{(j)}) = \sum_{k=1}^{|\mathcal{O}|} \alpha_k^{(i, j)} O_k^{(i,j)}(x^{(j)})
\end{equation}
where $\alpha^{(i, j)}$ is a differntiable $|\mathcal{O}|$-dim vector with its $k$-th elements associated with operation $O^k \in \mathcal{O}$ for the edge from $j$-th node to $i$-th node. $\alpha_k^{(i, j)}$ is normalized by the ``Gumbel-Softmax'' with reparameterization trick, for the purpose of efficient search process (see supplementary materials). $\mathcal{O}$ is the set of candidate operations, which contains six prevalent operations as below:
\vspace{-8pt}
\begin{multicols}{2}
\begin{itemize}
    \item $1 \times 1$ convolution 
    \item $3 \times 3$ convolution 
    \item $5 \times 5$ convolution 
    \item $3 \times 3$ average pooling 
    \item $3 \times 3$ max pooling 
    \item skip connection 
\end{itemize}
\vspace{-8pt}
\end{multicols}

\vspace{-8pt}
\subsection{Bi-level Optimization}
\vspace{-2pt}
To enable the optimization over network-level search space, one ad-hoc idea may be also adopting a continuously relaxed network-level search space and applying gradient-based updates, similar to the approach proposed by Liu \etal \cite{liu2019auto}. However, this continuous relaxation cannot support the need for multiple scales. As demonstrated by the single-branch network in Auto-Deeplab, the gradient-based search method can only pick the top1 single branch, which is not naturally applicable to discovering parallel branches. 

We instead unify the optimization of the network-level and cell-level architectures, and train a policy to control the selection of cross-scale connections. Specifically, we update the cell-level micro architecture via gradient descent, and our policy controller will sample the macro architecture of scales, where the network will activate the corresponding branches. The network-level macro search and cell-level micro search are integrated and alternatively proceed (see Sec. \ref{sec:search_detail} for searching details). The search procedure is shown in the $search$ function in Algorithm \ref{algo:darts}.
\vspace{-4pt}

\begin{algorithm}[tbh]
{\small
\SetAlgoLined
  \DontPrintSemicolon
  \SetKwFunction{CU}{cell\_update}
  \SetKwFunction{CTRLU}{controller\_update}
  \SetKwFunction{S}{search}
  \SetKwProg{Fn}{Function}{:}{}
  
    \Fn{\S{$N_{epoch}$,$N_{iter}$,$N_{ctr}$}}{
  initialize network weights $\omega$, cell architecture $\alpha$, branch architecture $\beta$, policy controller $\pi (\theta)$\;
  $epoch = 0$ \;
  \While{$epoch < N_{epoch}$}{
  $iter = 0$ \;
  \While{$iter < N_{iter}$}{
  $\beta\gets\pi(\theta)$\;
  \CU{$\beta$}\;
  $iter = iter + 1$
  }
  \CTRLU{$N_{ctr}$}\;
$epoch = epoch + 1$
  }
  }\;
  
  \Fn{\CU{$\beta$}}{
    Continuously relax operation $\Bar{o}^{(i,j)}$ weighted by the architecture parameter $\alpha^{(i,j)}$\\

    1. Update weights $\omega$ by descending $\nabla_{\omega}\mathcal{L}_{train}(\omega, \alpha, \beta)$\;
    2. Update cell architecture $\alpha$ by descending $\nabla_{\alpha}\mathcal{L}_{val}(\omega, \alpha, \beta)$\;

    3. Replace $\Bar{o}^{(i,j)}$ with $o_{k^*}^{(i,j)}, k^* = \mathrm{argmax}_{1\leq k \leq |\mathcal{O}|} \alpha_k^{(i,j)}$ \;
  }\;
  
  \Fn{\CTRLU{$N_{ctr}$}}{
    Fix network weights $\omega$ and cell architecture $\alpha$\;
    $step = 0$ \;
    \While{$step < N_{ctr}$}{
    Update controller by maximize the expected reward $\mathbb{E}_{\beta \sim \pi(\theta)}[\mathcal{R}(\omega, \alpha, \beta)]$\;
    $step = step + 1$
    }
  }

}
 \caption{Bi-level Optimization}\label{algo:darts}
\end{algorithm}

\vspace{-4pt}
\subsubsection{Network-level Macro Search}
\vspace{-4pt}
For the network-level macro search, we construct an RNN controller optimized via the REINFORCE gradient \cite{williams1992simple}, to decide which branches should be activated. The optimization of controller is illustrated in $controller\_update$ function in Algo. \ref{algo:darts}. At each time step, the controller will sample a $B$-dimension vector from the policy $\pi(\theta)$, where the $b$-th element indicates the activation status of the branch of scale $2^{b+1}$. We use average precision (AP) score on validation dataset as the reward $\mathcal{R}$. 

\vspace{-4pt}
\subsubsection{Cell-level Micro Search}
\vspace{-4pt}
With continuous relaxation mentioned in Sec. \ref{sec:cell}, we are able to optimize the discrete architecture with gradient descent. We adopt the first-order approximation in \cite{liu2018darts}. The loss function $\mathcal{L}$ is the mean squared error calculated between the regressed heatmap and the ground truth. The search algorithm proceeds the update of supernet weights $\omega$ and the cell-level architecture $\alpha$ in an alternative way, shown in function $cell\_update$ in Algo. \ref{algo:darts}.

\vspace{-2pt}
\subsection{Architecture Derivation}
\vspace{-2pt}
\paragraph{Network-level Architecture}
We adopt the optimized policy controller to sample the scale vector of each stage for $N$ times, then evaluate each status of parallel branches on the validation set by using the same shared weights in supernet. The one with the best performance will be selected as the discovered network-level architecture. We choose $N = 10$ by default, and a larger $N$ can be adopted to sample more status of parallel branches.

\vspace{-4pt}
\paragraph{Cell-level Architecture}
The optimal operators in cell-level architecture are derived by taking the $\mathrm{argmax}$ over $\alpha^{(i, j)}$. Different from previous works \cite{liu2019auto,liu2018darts}, we preserve the connection from all predecessors for every node.

\vspace{-4pt}
\section{Experiments}
\vspace{-4pt}
In all experiments, we use Nvidia Titan RTX for benchmarking the computing power. All experiments are performed under CUDA 9.0 and CUDNN V7, using Pytorch.\vspace{-4pt}

\subsection{Datasets}
\vspace{-2pt}
\paragraph{COCO Dataset}
We adopt the COCO dataset \cite{lin2014microsoft} as our main testbed, which contains more than 200k images and 250k person instances with 17 keypoints per instance. Our models are trained on the training dataset only, including 56k images and 150k person instances. During training, each image is resized to $256 \times 192$. We apply random flipping and rotation ([$-45 ^{\circ}$, $45 ^{\circ}$]) as data augmentation. The validation set contains about 5k images. Results are reported using the official evaluation metric of the COCO Keypoint Challenge: average precision (AP) and average recall (AR), based on object keypoints similarity (OKS):\vspace{-4pt}
\begin{equation} 
\begin{aligned}
\text{OKS} = \frac{\sum_i[\exp(-d_i^2/2s^2k_i^2)\delta(v_i>0)]}{\sum_i[\delta(v_i>0)]}
\end{aligned}
\end{equation}
Here, $d_i$ is the Euclidean distance between ground-truth and detected keypoint, $v_i$ is the visibility flag of the ground-truth keypoint. Each kind of keypoint also has a scale $s$ which is defined as the square root of the object segment area. $k_i$ is a predefined constant provided by the COCO Keypoint Challenge 2017 which is used to control falloff. \vspace{-8pt}

\paragraph{MPII Dataset}
The MPII dataset \cite{andriluka14cvpr} contains around 25k images with over 40k people with annotated body joints. Note that we use MPII to study the transferability of COCO-search architectures in this paper, rather than conducting any search on MPII directly. We trained our discovered model on training split only, and evaluate the single person pose estimation result on the validation dataset. We adopt the same data augmentation as on the COCO dataset. The input image size is set to $256 \times 256$. The evaluation metric is the Percentage of Correct Keypoints (PCK) \cite{yang2012articulated}.

\subsection{Architecture Search and Implementations}

\paragraph{Stem and Head Structure}\label{sec:stem_head}
In both our supernet and derived architecture, from the input image we leverage a pre-defined stem module to extract feature maps of different scales serving as inputs into the multi-scale modules. We design the ``stem'' structure with two $3\times 3$ convolutions (with stride 2) followed by a residual bottleneck \cite{he2016deep}, which produces a feature map of $4\times$ scale. We sequentially stack more $3\times 3$ convolutions with stride 2 to provide feature maps of higher scales. At the end of the network, the outputs of different branches are upsampled via nearest-neighbor interpolation to the highest resolution (``$4\times$'' branch in our case), and are aggregated with a concatenation. The concatenated feature map is then processed by a $3\times 3$ convolution of stride 1 towards the final output.
\vspace{-8pt}
\paragraph{Searching Details}\label{sec:search_detail}
 The number of branches $B$ is set to 4, where the scales of branches are $\{4 \times, 8\times, 16\times, 32 \times\}$ and the number of channels are 16, 32, 64, 128, respectively. The number of nodes in each individual cell is set as $K_I = 4$, while for the aggregation cell $K_M = 2$. Each stage has 2 multi-scale modules, where all stages will share the same cell-level architecture parameters $\alpha^{(i,j)}$, following previous works \cite{liu2019auto,liu2018darts,zoph2018learning}. We adopt Adam optimizer to train the cell-level architecture parameters $\alpha^{(i,j)}$ with a learning rate of $3\mathrm{e}{-3}$. During our search we set the total number of epochs $N_{epoch} = 50$, and $N_{iter} = 1100$ iterations per epoch. At the end of each epoch for the cell-level search, the scale controller will be updated for $N_{ctr} = 10$ times, with a batch size of 2. The controller is also optimized by an Adam optimizer, with a learning rate of $3.5\mathrm{e}{-3}$. For the reward to our controller, we use the OKS-based AP score on the validation dataset. Note that we start training the controller at the 5-th epochs, to avoid the bias caused by the under-trained supernet.
\vspace{-8pt}
\paragraph{Training Details}
The number of channels in the derived architecture is doubled from the supernet used in the search. The network is trained for 210 epochs. The initial learning rate is set to 0.001, which will be multiplied by a decay factor of 0.1 at epoch 170 and epoch 200. Adam optimizer is adopted for the network, with $\beta_1 = 0.9$, $\beta_2 = 0.99$. The batch size is set to 128.
\vspace{-8pt}
\paragraph{Testing Details}
We borrow the person detector from HTC \cite{chen2019hybrid} to get the person bounding box for COCO dataset testing. We follow the test procedure in \cite{newell2016stacked}, where the output heatmap is computed by averaging the output heatmap of the original and its flipped one.
\vspace{-2pt}

\subsection{Results}
\begin{figure*}[h]
\begin{center}
 \includegraphics[width=0.85\linewidth]{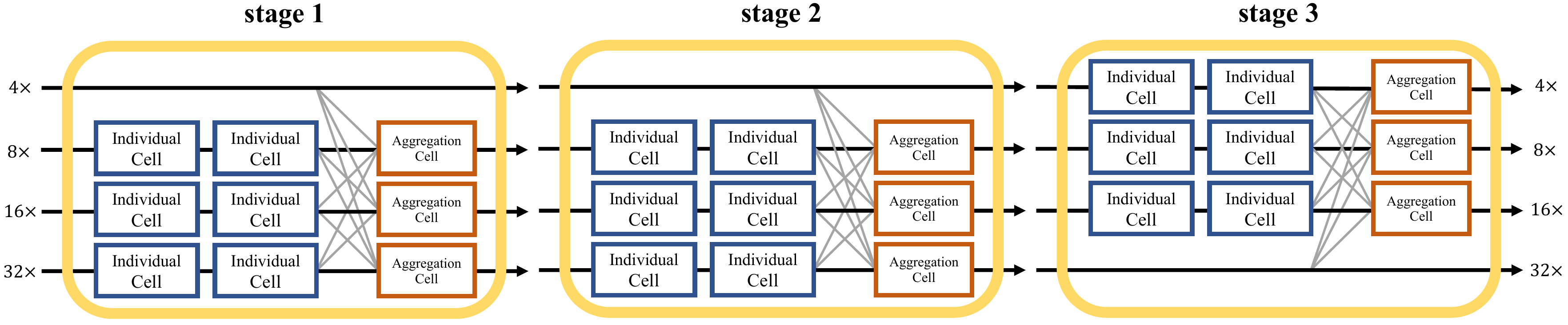}
\end{center}
\vspace{-4pt}
   \caption{
   The discovered network-level macro architecture of our proposed AutoPose framework.}
\label{fig:best_arch}
\vspace{-8pt}
\end{figure*}

\vspace{-2pt}
Fig. \ref{fig:best_arch} visualizes the best network-level architecture discovered by our AutoPose framework. Diversified scales are preferred at the beginning. Branches with different scales receive balanced activations, with feature maps of lower resolutions ($8\times,16\times, 32 \times$) are processed earlier and features of the highest resolution ($4\times$) are later processed. Please see our supplementary materials for the details of the searched cell structures.

\vspace{-8pt}
\paragraph{Results on COCO}
Tab. \ref{tab:coco} shows the performance of the discovered architecture on COCO validation dataset, compared with other state-of-the-art methods. Without any pretraining, our AP score significantly outperforms 8-stage Hourglass \cite{newell2016stacked} by 5.6 and CPN \cite{chen2018cascaded} by 5, with the same input size. We surpass the SimpleBaseline \cite{xiao2018simple} by 1.6 with 32.1\% less GFLOPs. Compared with the state-of-the-art HRNets, our model achieves competitive AP scores. AutoPose also shows clearly superior performance over the previous NAS for pose estimation work, PoseNFS \cite{yang2019pose}, with lower input image resolution and FLOPs.

\begin{table*}[h]
\caption{Comparison on the COCO validation set. The symbol $\dagger$ indicates the result obtained by reproducing with the author's official released code, as the original paper did not report the same setting. Models in the first block at table are hand-crafted models, while models in the second block are obtained by neural architecture search.}
\vspace{2pt}
\centering
{\small
\resizebox{0.8\linewidth}{!}{
    \begin{tabular}{ l || c | c | c | c c c c c c}
    \hline
    Method & Pretrain  & Image size & GFLOPs & AP & $\text{AP}^{50}$ & $\text{AP}^{75}$ & $\text{AP}^{M}$ & $\text{AP}^{L}$ & $\text{AR}$ \\\hline 
    8-stage Hourglass\cite{newell2016stacked} & N  & $256 \times 192$ & 14.30 & 66.9 & - & - & - & - & -\\ 
    8-stage Hourglass\cite{newell2016stacked} & N  & $256 \times 256$ & - & 67.1 & - & - & - & - & -\\ 

    CPN\cite{chen2018cascaded} & Y  & $256 \times 192$ & 6.20 & 68.6 & - & - & - & - & -\\ 
    Posefix\cite{moon2019posefix} & Y  & $256 \times 192$ & - & 71.5 & 88.0 & 77.6 & 68.0 & 78.1 & -\\ 
    SimpleBaseline-50\cite{xiao2018simple} & Y & $256 \times 192$ & 8.90 & 70.4 & 88.6 & 78.3 & 67.1 & 77.2 & 76.3\\ 
        SimpleBaseline-101\cite{xiao2018simple} & Y & $256 \times 192$ & 12.40 & 71.4 & 89.3 & 79.3 & 68.1 & 78.1 & 77.1\\ 
    SimpleBaseline-152\cite{xiao2018simple} & Y & $256 \times 192$ & 15.70 & 72.0 & 89.3 & 79.8 & 68.7 & 78.9 & 77.8\\ 
    HRNet-W32\cite{sun2019deep} & N & $256 \times 192$ & 7.10 & 73.4 & 89.5 & 80.7 & 70.2 & 80.1 & 78.9\\ 
     HRNet-W48\cite{sun2019deep} $\dagger$ & N & $256 \times 192$ & 14.60 & 73.5 & 89.5 & 80.6 & 70.4 & 79.8 & 79.2\\ \hline
     PoseNFS-3 \cite{yang2019pose} & Y & $384 \times 288$ & 14.8 & 73.0 & - & - & - & - & -  \\
    AutoPose (Ours) & N  & $256 \times 192$ & 10.65 & \textbf{73.6} & 90.6 & 80.1 & 69.8 & 79.7 & 78.1\\ 
        \hline
    \end{tabular}
    }
}
\label{tab:coco}
\vspace{-6pt}
\end{table*}

\vspace{-8pt}
\paragraph{Results on MPII}
We directly transfer the discovered architecture on the COCO dataset to train on the MPII dataset. We note that images from the MPII dataset are larger ($256\times 256$) than those in COCO ($256\times 192$), which is to our discovered architecture's disadvantage. Without any pretraining, the architecture designed by AutoPose surpasses ``ASR+AHO'' \cite{peng2018jointly}. We found that recent works (SimpleBaseline \cite{xiao2018simple} and HRNet \cite{sun2019deep}) adopt ImageNet pretraining and enjoy improved performance. Further discussion with the authors of HRNet \cite{sun2019deep} helped us confirm that adopting a pretraining backbone is vital to the performance boost and convergence for complicated pose estimation network, especially on relatively small datasets like MPII. 

This reveals a challenging issue when extending NAS to advanced computer vision tasks: as automatically designed architectures will require extra efforts to be pretrained on ImageNet, training from scratch is an unfair setting to NAS when compared with human-designed networks which adopt mature backbones pre-trained on extra data (e.g. ResNet101 \cite{he2016deep}). One remedy might be still leveraging a pretrained hand-crafted backbone followed by searched aggregation  \cite{chen2018searching,ghiasi2019fpn}; but that will limit the flexibility of NAS and somehow contradict to the philosophy of AutoML. We leave this open challenge for future work.


\subsection{Ablation Study}
\label{sec:eval}

\vspace{-4pt}
\paragraph{Multi-Scale Search Space}
Recently, Liu \etal proposed Auto-Deeplab \cite{liu2019auto} to search for both the cell-level and network-level architecture. However, their search architecture only allowed one scale to be activated per layer.

\begin{table}[h]
\caption{Transferability results on MPII validation dataset. No multi-scale testing technique is applied on our result.}
\centering
\resizebox{\linewidth}{!}{
    \begin{tabular}{ l || c | c c c c c c c | c}
    \hline
    Method & Pretrain & Head & Shoulder & Elbow & Wrist & Hip & Knee & Ankle & Total \\\hline 
    ``ASR+AHO'' \cite{peng2018jointly} & Y &97.3 &95.1 &88.7 &84.7 &88.4 &82.5 &78.1 &87.8 \\
    Hourglass\cite{newell2016stacked} & N &96.5 & 96.0 & 90.3 & 85.4 & 88.8 & 85.0 & 81.9 & 89.2 \\
    SimpleBaseline\cite{xiao2018simple} & Y & 97.0 & 95.9 & 90.3 & 85.0 & 89.2 & 85.3 & 81.3 & 89.6 \\
    HRNet-W32\cite{sun2019deep} & Y & 97.1 & 95.9 & 90.3 & 86.4 & 89.1 & 87.1 & 83.3 & 90.3  \\ \hline
    AutoPose (Ours) & N & 96.6 & 95.0 & 88.3 & 83.2 & 87.2 & 82.8 & 78.9 & 88.0  \\
        \hline
    \end{tabular}
    }
\label{tab:mpii}
\vspace{-5pt}
\end{table}

\begin{table}
\vspace{-1pt}
\caption{Evaluating our multi-scale  parallel branches and aggregation cells on the COCO validation set.}
\centering
\resizebox{1.0\linewidth}{!}{
    \begin{tabular}{ c | c || c c c c c c}
    \hline
    Aggregation Cell & Multi-branch & AP & $\text{AP}^{50}$ & $\text{AP}^{75}$ & $\text{AP}^{M}$ & $\text{AP}^{L}$ & $\text{AR}$ \\ \hline 
    searched & searched (single) & 63.9 & 86.5 & 70.2 & 61.2 & 68.9 & 69.0 \\
    N/A & searched & 70.8 & 90.4 & 77.6 & 67.2 & 77.0 & 75.3 \\
    searched & fixed to four & 71.8 & 89.5 & 79.4 & 69.6 & 75.5 & 74.8 \\
    searched & searched & 73.6 & 90.6 & 80.1 & 69.8 & 79.7 & 78.1\\
        \hline
    \end{tabular}
    }
\label{tab:ablation}
\vspace{-7pt}
\end{table}

\begin{figure*}[t!]
\vspace{-2pt}
\begin{center}
\includegraphics[width=0.88\linewidth]{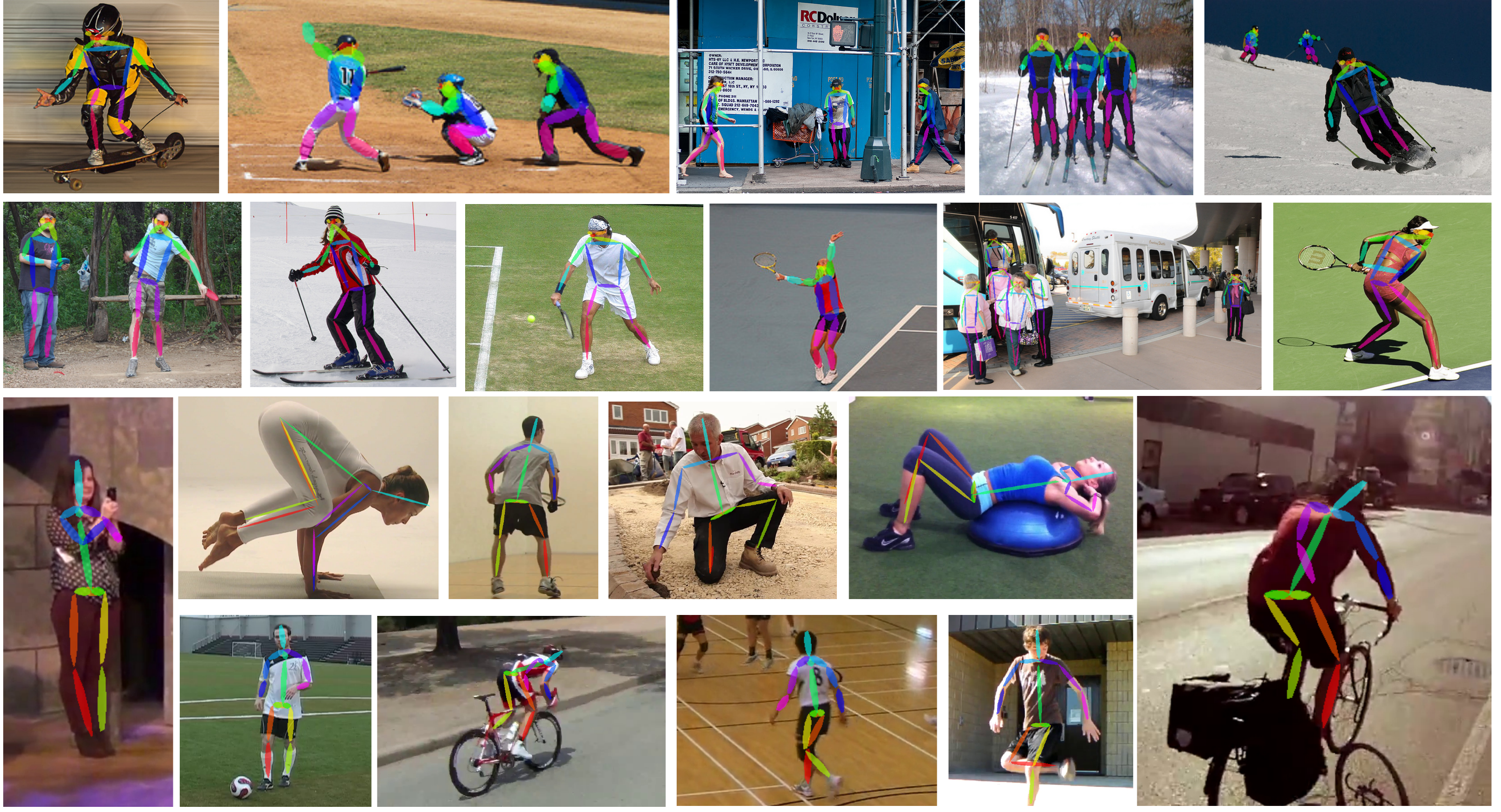} 
\caption{The visualizations of predicted keypoints. The first and last two rows are the results on COCO and MPII, respectively.}
\label{fig:qual}
\end{center}
\vspace{-8pt}
\end{figure*}

To demonstrate the effectiveness of the proposed multi-scale search space, we construct two variants of our network. First, we limit the maximum number of activated branch in each stage to one during search, i.e., degrading our parallel branches back to a single chain-like backbone. Second, we allow our multi-scale branches, but remove the context aggregations in our aggregation cells and instead adopt plain concatenation operations. Third, we manually keep all branches in all stages to be activated, and only search for the cell-level architecture.

The ablation results are shown in Tab. \ref{tab:ablation}.
When we only search one single scale per stage, the AP score dramatically drops to 63.9\%, demonstrating the key contribution of maintaining multiple scales throughout the network. When we activate multiple scales but bypass our aggregation cells, although the average precision under OKS = 0.5 achieves the similar bar (90.4\% v.s. 90.6\%), large OKS (0.75) will drop, implying the cross-scale context aggregation to be vital here. When we manually activate all four branches across all stages during search and evaluation, and only search for the micro cell-level structure, the performance remains to be lower than our proposed method.



\begin{figure}[h]
\begin{center}
 \includegraphics[width=0.85\linewidth]{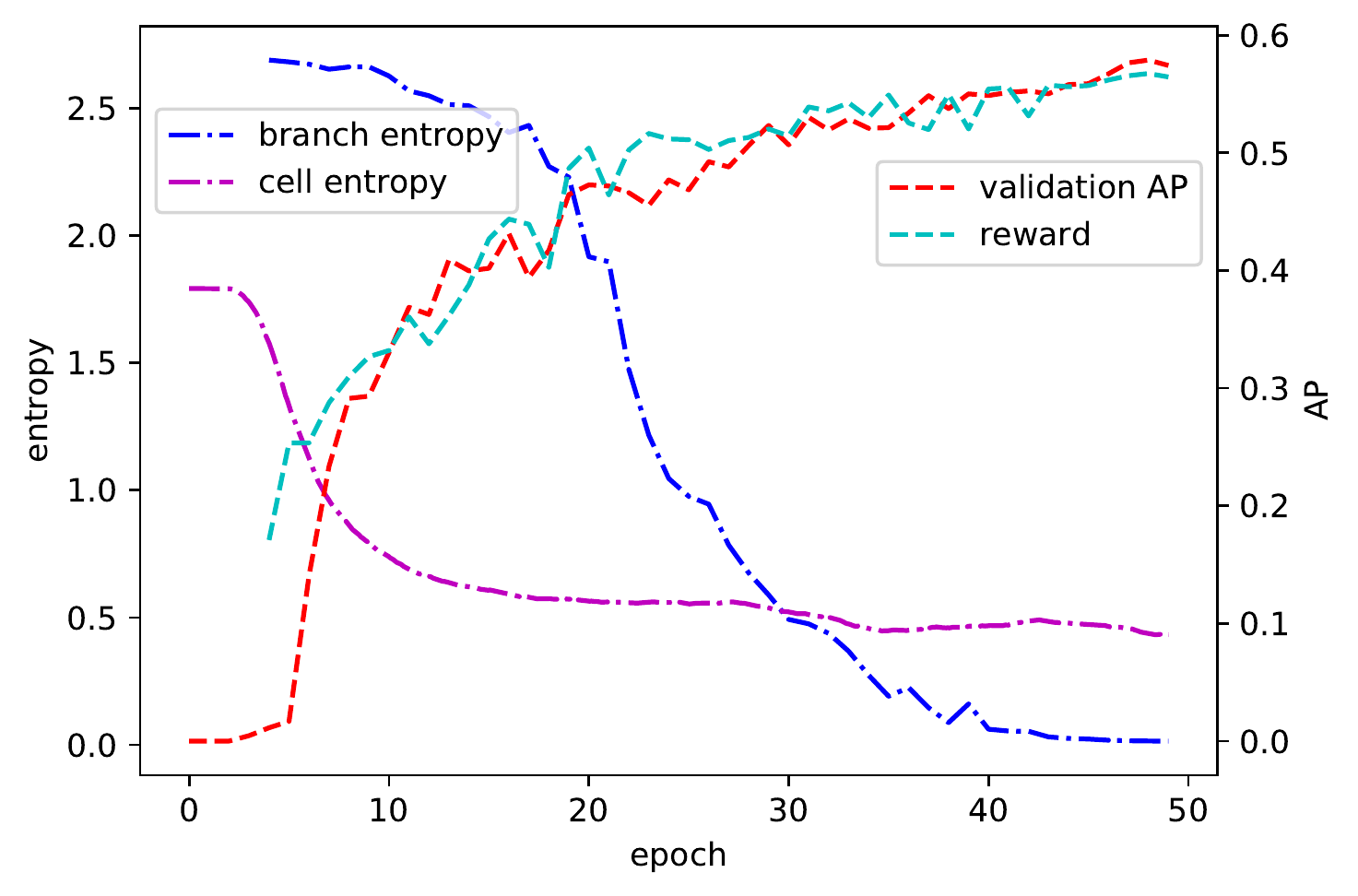}
\end{center}
\vspace{-1em}
   \caption{During the search process, the entropies of both cell-level and network-level architecture are decreasing (i.e. the supernet is becoming confident in its derived architecture), while the validation AP and controller's reward keep increasing (i.e. the performance of our supernet is being improved). Note that we begin to train the controller at 5-th epoch.}
\label{fig:entropy}
\vspace{-4pt}
\end{figure}

\vspace{-8pt}
\paragraph{Bi-level Optimization}
As existing NAS works only adopt a single search algorithm throughout their framework, we are the first to leverage a bi-level optimization for different aspects of our search space. Fig. \ref{fig:entropy} visualizes the convergences and performance during our search process.
We can see that the validation accuracy steadily improves throughout the search process, while both the cell-level and network-level optimizations are converged to a small subset of the search space, indicated by the decreasing entropies. \vspace{-4pt}

\vspace{-8pt}
\paragraph{Image Size during searching}
Previous work \cite{liu2019auto} reduced the input image size during the search phase to accelerate NAS. However, we argue that for dense image prediction tasks like pose estimation which requires fine details in context, adopting reduced input size during search amplifies the gap between the search and the train from scratch processes. We conduct an experiment where we use the size of $128\times96$ during searching, while still keep the size of $256\times192$ during training from scratch. In Tab. \ref{tab:size}, we can see an obvious performance drop when the image size is reduced during searching.

\begin{table}[h]
\caption{Impact of image sizes during search (evaluated on the COCO validation set).}
\centering
\resizebox{1.0\linewidth}{!}{
    \begin{tabular}{ c | c || c c c c c c}
    \hline
\multicolumn{2}{c||}{Image size} & \multirow{2}{*}{AP} & \multirow{2}{*}{$\text{AP}^{50}$} & \multirow{2}{*}{$\text{AP}^{75}$} & \multirow{2}{*}{$\text{AP}^{M}$} & \multirow{2}{*}{$\text{AP}^{L}$} & \multirow{2}{*}{$\text{AR}$} \\
\multicolumn{1}{c}{Search} & \multicolumn{1}{c||}{Train} &  &  &  &  &  &  \\ \hline
\multicolumn{1}{c|}{$128 \times 96$} & $256 \times 192$ & 70.1 & 89.6 & 75.9 & 66.1 & 75.8 & 74.5 \\
\multicolumn{1}{c|}{$256 \times 192$} & $256 \times 192$ & 73.6 & 90.6 & 80.1 & 69.8 & 79.7 & 78.1 \\
\hline
    \end{tabular}
    }
\label{tab:size}
\vspace{-2pt}
\end{table}

\vspace{-6pt}
\section{Conclusion}
\vspace{-4pt}
We propose AutoPose for the 2D human pose estimation task. Our high-capacity multi-scale search space supports the discovery of multiple parallel branches of diverse scales. By leveraging both gradient-based cell-level search and reinforcement learning based network-level search, our AutoPose successfully discovers the architecture that achieves very competitive performance on COCO and MPII dataset. We believe that pursuing an optimal combination of both cell-level and network-level structure is vital to accurate pose estimation, and wish our work to further advocate more NAS practice in advanced computer vision applications, with more reflections in the pros and cons.

\clearpage
{\small
\bibliographystyle{ieee_fullname}
\bibliography{egbib}
}

\end{document}